SPIE Proceedings Publications# Emerging Synergies Between Large Language Models and Machine Learning in E-commerce Recommendations

Xiaonan Xu[1]*, Yichao Wu[2], Penghao Liang[3], Yuhang He[4], Han Wang[5]
[1]Independent Researcher, Northern Arizona University, Flagstaff,
[2]Computer Science, Northeastern University, Boston, MA, USA
[3]Information Systems, Northeastern University, Boston, MA, USA
[4] Computer Science and Technology, Tianjin University of Technology, Tianjin, China
[5]Financial Mathematics, University of Southern California, Los Angeles, USA
*Corresponding author: Xiaonan Xu, [E-mail: xiaonanxu5@gmail.com]**ABSTRACT:** With the boom of e-commerce and web applications, recommender systems have become an important part of our daily lives, providing personalized recommendations based on the user's preferences. Although deep neural networks (DNNs) have made significant progress in improving recommendation systems by simulating the interaction between users and items and incorporating their textual information, these DNN-based approaches still have some limitations, such as the difficulty of effectively understanding users' interests and capturing textual information. It is not possible to generalize to different seen/unseen recommendation scenarios and reason about their predictions. At the same time, the emergence of large language models (LLMs), represented by ChatGPT and GPT-4, has revolutionized the fields of natural language processing (NLP) and artificial intelligence (AI) due to their superior capabilities in the basic tasks of language understanding and generation, and their impressive generalization and reasoning capabilities. As a result, recent research has sought to harness the power of LLM to improve recommendation systems. Given the rapid development of this research direction in the field of recommendation systems, there is an urgent need for a systematic review of existing LLM-driven recommendation systems for researchers and practitioners in related fields to gain insight into. More specifically, we first introduced a representative approach to learning user and item representations using LLM as a feature encoder. We then reviewed the latest advances in LLMs techniques for collaborative filtering enhanced recommendation systems from the three paradigms of pre-training, fine-tuning, and prompting. Finally, we had a comprehensive discussion on the future direction of this emerging field.

**Keyword list:** Large language model; Deep learning; E-commerce personalized recommendation; Collaborative filtering algorithm## 1. INTRODUCTION

In recent years, large language models (LLMs) have made significant progress in natural language processing research. Although a great deal of work has been done on developing universal LLMs, less research has been done on the potential of LLMs in recommendation systems. In this paper, we propose a new framework, PALR, which aims to combine historical user behavior (e.g., clicks, purchases, ratings, etc.) with LLM to generate items of user preference. Specifically, we first utilize user/item interactions as a guide for candidate retrieval, and then employ an LLM-based ranking model to generate recommended items. Unlike existing methods, we not only fine-tune the universal LLM for zero sample/small sample tests but also fine-tune the LLM on the 700 billion parameter scale for ranking. The model takes the search candidates as inputs in a natural language format and explicitly requires the selection of results from the input candidates when reasoning. Our experimental results show that our solution outperforms the latest technology in a variety of sequential recommendation tasks.

A recommendation system is an information filtering system designed to predict and recommend items or products. These systems are widely used in the e-commerce, online advertising, social media, and entertainment industries. In recent years, the emergence of large language models (LLMs), such as BERT, GPT-3, FLAN-T5, etc., has brought major breakthroughs in natural language processing research. Encouraged by these advances, researchers began to explore the potential of using LLMs in recommendation systems.

LLMs have several unique advantages over traditional recommendation modeling techniques and more recent sequential modeling and graph modeling techniques. First, LLM essentially supports perceptual learning and does not require pre-training embeddings for each project. Instead, each item can be represented as a piece of text. Second, LLM can easily integrate various signals such as metadata, context, and multimodal signals into the recommendation process by incorporating them into the model prompts. Third, LLM can transfer knowledge gained from one domain to another, providing significant advantages in cold start scenarios where user behavior data is limited. Finally, LLMs have tremendous knowledge and excellent reasoning ability, as well as the ability to generate natural language output. As a result, they can provide understandable explanations of recommendations, enhancing user trust and engagement.

Intelligent recommendation service is one of the important technical means to improve the sales conversion rate of e-commerce websites. It has important differences from traditional search technology. Intelligent recommendation service can provide information more accurately, save users' time in searching for information, and improve the accuracy of searching for information. Therefore, this paper combines personalized recommendation based on large language model with collaborative filtering recommendation algorithm to realize the actual personalized recommendation process of e-commerce platform and analyzes the development prospect of artificial intelligence in the field of e-commerce according to the combination of collaborative filtering algorithm and large language model.

## 2. THEORETICAL OVERVIEW

### 2.1 AI drives intelligent recommendation

With the rapid development of the Internet, the e-commerce industry has become the main channel for consumers to shop. How to let consumers quickly and accurately find the goods they need is a problem that the e-commerce industry has been exploring. In recent years, with the gradual application of artificial intelligence technology, a new model has emerged in the e-commerce industry - intelligent recommendation.

Intelligent recommendation is the use of artificial intelligence technology, through the analysis of users' historical behavior, to recommend products that users are interested in. This recommendation can not only help consumers quickly find the goods they need but also improve the sales of e-commerce platforms. Behind the intelligent recommendation are some common algorithm models, such as collaborative filtering, deep learning, etc., which can predict users' shopping preferences based on their historical behavior and make personalized recommendations.

Although intelligent recommendation has been gradually popularized in the e-commerce industry, there are still some problems. For example, if the algorithm model is not trained well, then the recommended goods may not match the actual needs of the user. Moreover, for some consumers, they may feel that intelligent recommendations are not accurate and limited because the algorithm model may only recommend based on the historical behavior of the user, without taking into account the actual needs of the user. In addition, some consumers may not like smart recommendations because they want to be more independent in their shopping decisions.

## 2.2 Recommendation System (RecSys)

Recommendation systems (RecSys) play an important role in reducing information overload and enriching the online experience for users who need to filter through large amounts of information to find what interests them. In several applications such as entertainment, e-commerce, and job matching, the recommendation system provides personalized candidate recommendations based on user preferences. For example, movie recommendation systems, such as IMDB and Netflix, recommend the latest movies based on the content of the movie and the user's past interaction history to help users discover new movies that fit their interests. The basic idea of the recommendation system is to use the interaction between the user and the item and its related side information, especially text information (such as the title or description of the item, the user profile, and the user's comments on the item), to predict the match score between the user and the item (that is, the probability that the user likes the item). More specifically, collaborative behavior between users and objects has been used to design various recommendation models that can also be further used to learn representations of users and objects. In addition, text information about users and items contains a wealth of knowledge that can help calculate matching scores, providing a huge opportunity to understand user preferences and thus drive the development of recommendation systems.

Deep neural network (DNN) has been widely used to promote the development of recommendation system because of its outstanding ability of representation learning in various fields. For example, recurrent neural networks (RNNs), which are particularly effective tools for sequence data, have been used to capture higher-order dependencies in user interaction sequences. Considering that users' online behaviors (such as chicks, purchases, socializing) are graph-structured data, graph neural networks (GNNs) have become an advanced representation learning technique for learning user and item representations. At the same time, DNNs have demonstrated a unique ability to model user and project interactions. At the same time, DNN also shows advantages in encoding side information. For example, a Bert-based approach has been proposed to extract and exploit users' text comments.

Despite these achievements, most existing advanced recommendation systems still face some inherent limitations. First, due to the limitations of model size and data size, DNN-based models (such as CNN and LSTM) and pre-trained language models (such as BERT) previously used for recommendation systems are unable to adequately capture the text knowledge of users and projects, which indicates their poor natural language understanding. As a result, the prediction performance in various recommendation scenarios is not ideal.

## 2.3 Collaborative Filtering(CF)

The recommendation system has become a basic service of the Internet, which recommends personalized products to users by learning their preferences in historical interaction behaviors. At present, collaborative filtering algorithms based on Graph Neural Networks have shown great advantages in the field of recommendation. Generally speaking, in the Collaborative Filtering (CF) scenario, we have a user set U and item set I, and their interactions. So, if we treat each user and item as a node and the interaction record between them as an edge, we can construct a User-Item Interaction Graph. Then, based on the layer of information transmission and aggregation of graph neural networks, we can finally get the Representation learned by each user and commodity node based on the graph structure. Since this representation contains the information of collaborative filtering, we can call it CF-side representation.

Filter the vast amount of information together with the feedback, evaluation and opinions of similar users, and select the information that users may be interested in.

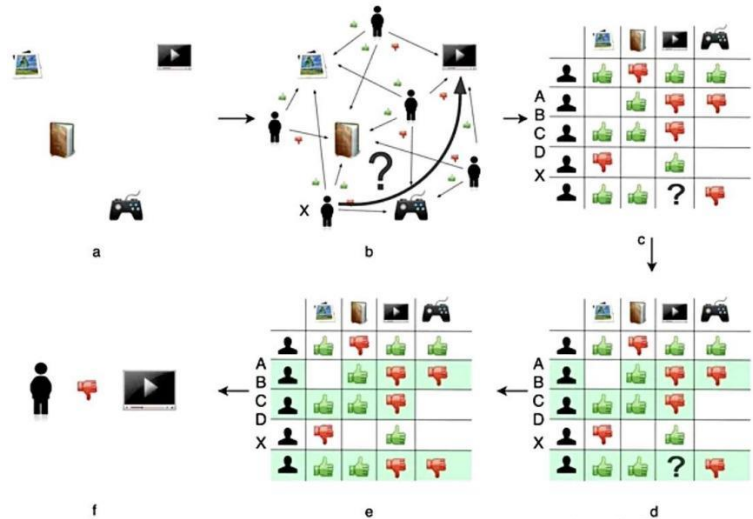

**Figure 1:** Framework of the basic principles of collaborative filtering

The evaluation of collaborative filtering algorithm recommendation systems is to assess whether a recommendation system is good. A good recommendation system can not only accurately predict the user's behavior but also expand the user's vision and help the user find things that they may be interested in but are not so easy to find, thus increasing the revenue benefit through the recommendation system. For example, if you predict that a user will buy a toothbrush in the future, the prediction is obviously accurate, but because the user does not need or does not choose your product when they need it, it will not increase your revenue, so it is not a good recommendation.

## 2.4 Basic principle of personality recommendation matrix decomposition

The disadvantage of collaborative filtering is that the co-occurrence matrix is generally sparse, and the process of finding similar users is not accurate in the case of few user behaviors. The matrix decomposition method improves the ability to deal with sparse matrices. Matrix decomposition is to use the co-occurrence matrix to generate the hidden vector of the item and the user. The hidden vector can be understood as the same vector space in the same vector similarity to sort recommended.

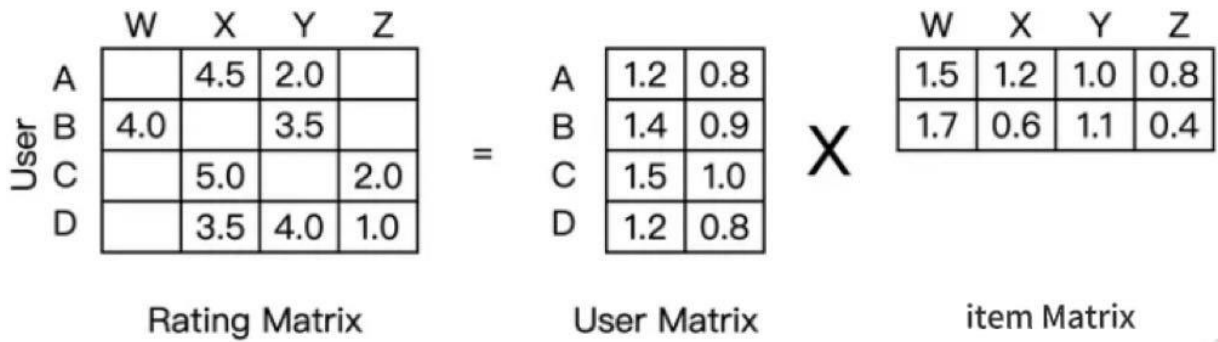

**Figure 2:** Matrix decomposition structure diagram

The process of generating implicit vectors is to imitate the process of matrix multiplication. The user implicit vector is the corresponding row vector of the user matrix, and the item implicit vector is the corresponding column vector of the item matrix. It is hoped that the product of the user matrix and the item matrix is as close as possible to the original co-occurrence matrix. The model iteration method is gradient descent. The implicit vector dimension k is a hyperparameter. However, this method cannot realize the user similarity value by itself, so it needs to realize the user personalized recommendation through the collaborative algorithm combined with the large language model.

In this section, we will delve into the integration of large language models (LLMs) within the framework of collaborative filtering (CF) for recommendation systems. Specifically, we will explore how LLMs can augment traditional CF approaches to enhance recommendation accuracy and personalize user experiences. This integration involves leveraging the natural language understanding and generation capabilities of LLMs to refine user-item interactions and enrich item representations.

## 3. METHODOLOGY

In the field of E-commerce, personalized recommendation systems are essential to improve user satisfaction and increase sales. In combination with the user-based collaborative filtering algorithm of the large language model, K users who are most similar to the target user are selected, items not yet purchased by the target user are selected from the order records of K users for recommendation, the recommendation score of these K users is calculated, and the N items with the highest score are selected to generate the final recommendation list. The main steps are as follows.

Collaborative filtering recommendation algorithm refers to the clustering of items with similar attributes or characteristics to recommend these similar items to users. This method is suitable for situations where the number of items is much larger than the number of users, such as recommending movies, music, books, etc. By dividing users into different groups, people can better understand the needs and interests of different users, and recommend items that the current user has not operated. This method is suitable for situations where the number of users is much larger than the number of items, such as social networks, shopping sites, etc.

The recommendation algorithm consists of three steps: collecting user preferences, finding similar users or items, and calculating recommendation.

Collect user preferences → Find similar items or users → Calculate recommendations

### 3.1 Calculate the similarity between users

# SPIE Proceedings Publications

(1) Build a list of users or items
Assume that four users A, B, C and D have scored five items a, b, c, d and e, and a user-item rating table can be created according to the score, as shown in Table 1:

Table 1: List of users and item types

| User | Type1 | Type2 | Type3 | type4 | Type5 |
|---|---|---|---|---|---|
| A | 3.0 | 4.0 | 0 | 3.5 | 0 |
| B | 4.0 | 0 | 4.5 | 0 | 3.5 |
| C | 0 | 3.5 | 0 | 0 | 3.0 |
| D | 0 | 4.0 | 0 | 3.5 | 3.0 |

(2) Calculation of similarity
Based on the collected data, algorithms (such as Pearson correlation coefficient, cosine similarity, etc.) are used to calculate the similarity between the target user and other users. By analyzing the language expression and content preferences of users, the large language model can provide a richer dimension of similarity calculation, thus helping to identify the user groups with truly similar purchasing behaviors and preferences.

First, we have to predict user 3's rating for item 4. In a user-based recommendation system, we find three users who are most similar to user 3 and use the ratings of these three users to predict user 3's rating of item 4.

Common similarity measures are cosine, Pearson, Euclid, and so on. We will use cosine similarity here, which is defined as follows:

$$\text{similarity} = \cos(\theta) = \frac{A \cdot B}{\|A\|\|B\|} = \frac{\sum_{i=1}^{n} A_i B_i}{\sqrt{\sum_{i=1}^{n} A_i^2} \sqrt{\sum_{i=1}^{n} B_i^2}} \quad (1)$$

Moreover, Pearson correlation is defined as:

$$r = \frac{\sum_{i=1}^{n}(x_i - \bar{x})(y_i - \bar{y})}{\sqrt{\sum_{i=1}^{n}(x_i - \bar{x})^2} \sqrt{\sum_{i=1}^{n}(y_i - \bar{y})^2}} \quad (2)$$

In sklearn, the Nearest Neighbors method can be used to search for k nearest neighbors based on various similarity measures.

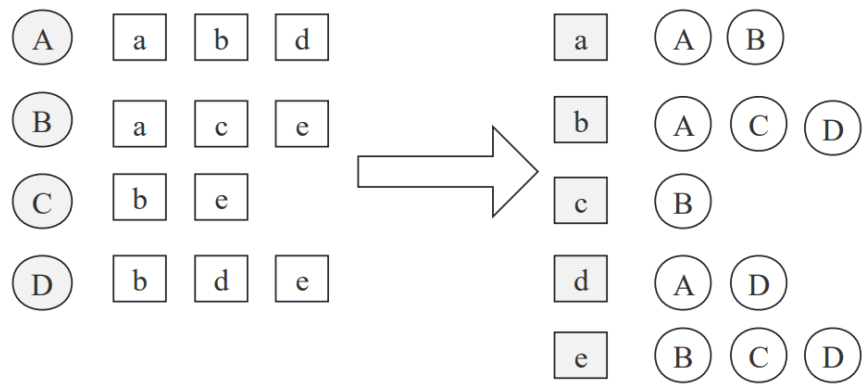

**Figure 3:** Similar user - item inversion list

The user-based recommendation system method further predicts user 3's rating for item 4 based on predict_user's function. The prediction is calculated by a weighted average of the deviations from the neighbor average and added to the average score of the target user. Bias is used to adjust for user-related bias. User bias occurs because some users may always give high or low ratings to all items.

$$p_{a,i} = \bar{r}_a + \frac{\sum_{u \in K}(r_{u,i} - \bar{r}_u) \times w_{a,u}}{\sum_{u \in K} w_{a,u}} \quad (3)$$

### 3.2 Item-Based collaborative filtering

Similar to the UserCF algorithm, a user-item inversion table is created when ItemCF is used to calculate item similarity. Then, for each user, each pair of items in his item list can be added by 1 to the co-occurrence matrix C

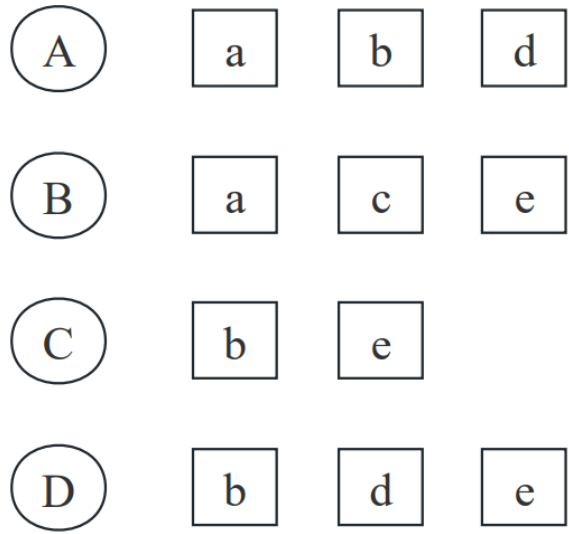

**Figure 4:** UserCF algorithm model

In this method, the cosine similarity measure is used to calculate the similarity between a pair of goods. You can predict target user a's rating of target item i by using a simple weighted average:

$$p_{a,i} = \frac{\sum_{j \in K} r_{a,j} w_{i,j}}{\sum_{j \in K} |w_{i,j}|} \quad (4)$$

To implement Adjusted Cosine similarity in Python, I defined a simple function called computeAdjCosSim, which returns the adjusted cosine similarity matrix, giving the score matrix. The functions findksimilaritems_adjcos and predict_itembased_adjcos use the adjusted cosine similarity to findksimilaritems and calculate the predicted score.

The recommend Item function prompts the user to select the recommendation method (based on user (Cosine), based on user (correlation), based on product (cosine), based on product (Adjusted Cosine Similarity). Based on the selected method and similarity measure, the function can predict the rating of the specified user and the item and suggest whether the item can be recommended to the user, if the item has not been rated by the user and the predicted rating is greater than 6, then recommended to the user, if the rating is less than 6, then not recommended to the user.

### 3.3 Evaluation of accuracy of recommendation algorithm

Python 3.9.0 is used to implement the above recommendation algorithm, and Precision and Recall are used to evaluate the accuracy of the algorithm. It is calculated as follows:

$$\text{Precision} = \frac{TP}{TP+FP} \times 100\% \qquad (5)$$

$$\text{Recall} = \frac{TP}{TP+FN} \times 100\% \qquad (6)$$

Where: TP is the number of positive samples predicted by the model to be positive; FP is the number of negative samples predicted by the model to be positive; FN is the number of positive samples predicted by the model to be negative. Precision and Recall are improved based on the product recommendation scenarios of e-commerce shopping platforms. Meanwhile, considering that enterprises prefer products to be recommended by a wider range, rather than just popular products, this paper introduces coverage Cover to evaluate the recommendation algorithm. Its calculation formula is:

$$\text{Cover} = \frac{\text{Number of all recommended product categories}}{\text{The number of categories of all products on the platform}} \times 100\% \qquad (7)$$

Based on the principle of collaborative filtering recommendation algorithm, its core idea is to realize personalized recommendation through the similarity between users or similarities between items. This method assumes that users' preferences for similar items have some consistency, so that users' historical behavior data can be used to make recommendation prediction. In this process, the role of the large language model is self-evident: it can process huge amounts of user and object data, and mine more refined similarity relationships from it. By comprehensively considering more dimensions of information such as text and semantics, the large language model can more accurately capture the association between user interests and item features, thus improving the accuracy and personalization of recommendations. Therefore, user-based collaborative filtering algorithms focus on calculating the similarity between users and making recommendations by finding common interests among users. The item-based collaborative filtering algorithm looks for similar items from the perspective of items and recommends them to users. In this process, large language models can provide richer feature representation and similarity calculation for collaborative filtering algorithms, thus improving the coverage and accuracy of recommendation systems. By combining collaborative filtering algorithms and large language models, the recommendation system can better understand the user's behavior and preferences, and achieve more accurate and personalized recommendation services.

## 4. EXPERIMENT AND RESULT

The performance of the collaborative filtering algorithm combined with the large language model is compared in the intelligent recommendation experiment of e-commerce products. The first group is the traditional UserCF algorithm and the improved T-UserCF algorithm. In the experiment, their performance is evaluated by comparing their MAE values. The second group is the traditional ItemCF algorithm and the improved T-ItemCF algorithm. In the experiment, MAE value is also used to evaluate their performance.

(1) The performance of UserCF algorithm and TUserCF algorithm is compared in the experiment. The number of neighbors K starts from 5 and gradually increases with 5 as the basic unit. Then, the average absolute error (MAE) values of these two algorithms are compared under different number of neighbors.

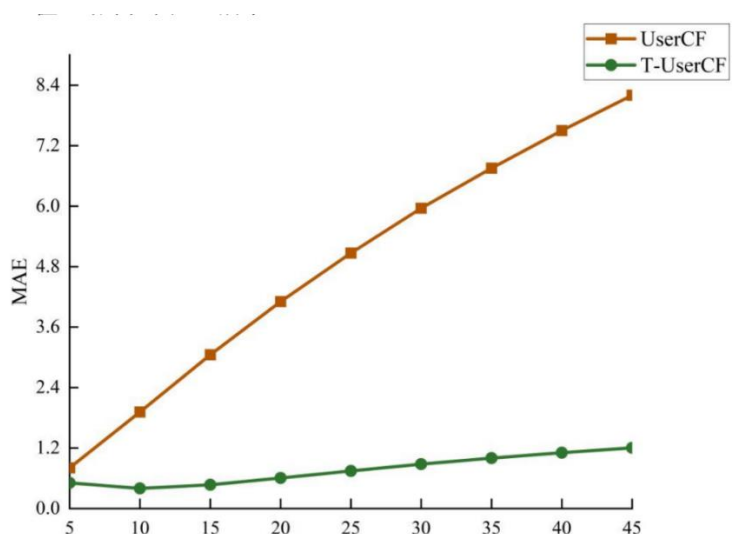

**Figure 4:** The MAE value of the UserCF algorithm and T-UserCF algorithm

When K value is 10, the MAE value of T-UserCF algorithm reaches the lowest point. When K is 5, the MAE value of UserCF algorithm reaches the lowest point and then continues to rise. According to the experimental results, it can be concluded that the accuracy of UserCF algorithm and TUserCF algorithm is increasing with the increasing of K value. When the number of neighbors K is the same, the TUserCF algorithm has better recommended performance than the UserCF algorithm.

**Table 2:** UserCF and T-UserCF recommend algorithm values

| Number of neighbors | NAME | |
| --- | --- | --- |
| | UserCF | T-UserCF |
| 5 | 0.807 | 0.508 |
| 10 | 1.915 | 0.398 |
| 15 | 3.050 | 0.471 |
| 20 | 4.107 | 0.605 |
| 25 | 5.070 | 0.746 |
| 30 | 5.960 | 0.880 |

Based on this experimental section, the following summary can be provided:
1. Performance of UserCF vs. T-UserCF Algorithm:
   - The experiment compared the traditional UserCF algorithm with an improved version called T-UserCF.
   - The Mean Absolute Error (MAE) values were used to evaluate the performance of these algorithms under varying numbers of neighbors (K).

   - It was observed that as the number of neighbors (K) increased, the accuracy of both algorithms improved.
   - Notably, when K was set to 10, the T-UserCF algorithm achieved the lowest MAE value, indicating higher accuracy compared to UserCF.
   - Overall, the T-UserCF algorithm demonstrated better recommendation performance across different values of K compared to the traditional UserCF algorithm.
2. Performance of ItemCF vs. T-ItemCF Algorithm:
   - Similarly, another comparison was conducted between the traditional ItemCF algorithm and an improved version called T-ItemCF.
   - MAE values were again utilized for performance evaluation.
   - Results showed that T-ItemCF outperformed ItemCF across different values of K, with the lowest MAE achieved at specific K values.
   - This suggests that the improvements incorporated into T-ItemCF led to enhanced recommendation accuracy compared to the traditional ItemCF algorithm.

Overall, these experiments underscore the effectiveness of the enhancements made in T-UserCF and T-ItemCF algorithms over their traditional counterparts. The integration of large language models has evidently contributed to refining the collaborative filtering process, resulting in more accurate and personalized recommendations in e-commerce product scenarios.

## 5. CONCLUSION

In the realm of E-commerce, personalized recommendation systems are pivotal for enhancing user satisfaction and driving sales. By combining user-based collaborative filtering with LLMs, this study selects similar users, retrieves items from their purchase history, and calculates recommendation scores based on LLM-guided features. Collaborative filtering algorithms, whether user-based or item-based, rely on similarity metrics to make recommendations. LLMs enrich these algorithms by providing more refined feature representations and enhancing similarity calculations, thereby improving recommendation accuracy and personalization. Experimental evaluations compare the performance of traditional collaborative filtering algorithms with their LLM-enhanced counterparts. Both user-based and item-based algorithms demonstrate improved accuracy and recommendation performance when augmented with LLMs. Notably, the T-UserCF and T-ItemCF algorithms exhibit superior performance over their traditional counterparts across various evaluation metrics. These findings underscore the efficacy of integrating LLMs with collaborative filtering techniques to achieve more accurate and personalized recommendations in e-commerce settings.

In conclusion, the integration of large language models with collaborative filtering algorithms holds immense promise for advancing recommendation systems in e-commerce. LLMs, with their superior language understanding and generation capabilities, enrich collaborative filtering processes by enabling more accurate user-item similarity calculations and personalized recommendations. By leveraging LLMs' capabilities, recommendation systems can overcome traditional limitations and deliver enhanced user experiences, ultimately driving sales and fostering user trust in e-commerce platforms.

**REFERENCE**

# SPIE Proceedings Publications

# SPIE Proceedings Publications